\documentclass[11pt,a4paper]{article}
\usepackage{tabu}
\usepackage{booktabs}
\usepackage{multirow}
\usepackage{subcaption}
\usepackage{fancyvrb}
\usepackage{graphicx}

\usepackage[hyperref]{emnlp2020}
\usepackage{times}
\usepackage{latexsym}

\usepackage{microtype}

\aclfinalcopy 
\def\aclpaperid{875} 

\title{The Multilingual Amazon Reviews Corpus}
\author{Phillip Keung\textsuperscript{•} \quad Yichao Lu\textsuperscript{•} \quad György Szarvas\textsuperscript{•} \quad Noah A. Smith\textsuperscript{†‡} \\
\textsuperscript{•}Amazon \quad \textsuperscript{†}University of Washington \quad \textsuperscript{‡}Allen Institute for AI \\
{\small \tt \{keung,yichaolu,szarvasg\}@amazon.com \quad nasmith@cs.washington.edu}
}

\date{}

\begin{document}
\maketitle
\begin{abstract}
We present the Multilingual Amazon Reviews Corpus (MARC), a large-scale collection of Amazon reviews for multilingual text classification. The corpus contains reviews in English, Japanese, German, French, Spanish, and Chinese, which were collected between 2015 and 2019. Each record in the dataset contains the review text, the review title, the star rating, an anonymized reviewer ID, an anonymized product ID, and the coarse-grained product category (e.g., `books', `appliances', etc.) The corpus is balanced across the 5 possible star ratings, so each rating constitutes 20\% of the reviews in each language. For each language, there are 200,000, 5,000, and 5,000 reviews in the training, development, and test sets, respectively. We report baseline results for supervised text classification and zero-shot cross-lingual transfer learning by fine-tuning a multilingual BERT model on reviews data. We propose the use of mean absolute error (MAE) instead of classification accuracy for this task, since MAE accounts for the ordinal nature of the ratings.
\end{abstract}

\section{Introduction}

\begin{table*}[h]
	\centering
	\footnotesize
\centering
\begin{tabu}{lcccccc}
\toprule
 & \textbf{En} & \textbf{De} & \textbf{Es} & \textbf{Fr} & \textbf{Ja} & \textbf{Zh} \\
\midrule 
Number of products & 196,745 & 189,148 & 179,076 & 183,345 & 185,436 & 164,540 \\
Number of reviewers & 185,541 & 171,620 & 150,938 & 157,922 & 164,776 & 132,246 \\
Average characters/review & 178.8 & 207.9 & 151.3 & 159.4 & 101.4 & 51.0 \\
Average characters/review title & 24.2 & 21.8 & 19.2 & 19.1 & 9.5 & 7.6 \\
\bottomrule
\end{tabu}
\caption{Training corpus statistics. We provide 200,000 reviews per language.}
\label{table:statistics}
\end{table*}

Text classification is one of the fundamental tasks in natural language processing, and research in this area has been accelerated by the abundance of corpora across different domains (e.g., Twitter sentiment \cite{pak2010twitter}, movie ratings \cite{imdb}, textual entailment \cite{snli:emnlp2015}, restaurant reviews \cite{yelp}, among many others).

The construction of \emph{multilingual} classification systems which handle inputs from different languages has been studied extensively in previous work (e.g., \citealp{bel2003cross, de2007multilingual}). More recently, researchers have observed `zero-shot' cross-lingual behavior \citep{lu2018neural, artetxe2019massively} where classification performance in one language can be transferred to the same task in another, without target language supervision, as long as the encoder was pretrained on a machine translation task. In addition, contextual embeddings have shown unexpected cross-lingual behavior in classification, NER, and dependency parsing tasks \citep{wu2019beto,keung2019adversarial,conneau2019unsupervised}.

\begin{figure}
\begin{Verbatim}[fontsize=\small]
{     
   "review_id": "en_0000258",
   "reviewer_id": "reviewer_en_0010355",
   "product_id": "product_en_0000097",
   "language": "en",
   "stars": 5,
   "review_title": "Salad Spinner",
   "review_body": "Perfect for herbs and 
               leafy vegetables!",
   "product_category": "kitchen"
}
\end{Verbatim}
\caption{A hypothetical review from our corpus.} 
\label{review-example}
\end{figure}

As with all other areas in NLP, progress in multilingual research relies on the availability of high-quality data. However, large-scale multilingual text classification datasets are surprisingly rare, and existing multilingual datasets have some notable deficiencies. 

The proprietary Reuters RCV1 \cite{lewis2004rcv1} and RCV2 \cite{rcv2} corpora and its derivatives like MLDoc \cite{mldoc} are relatively small; in RCV2, each language has ${\sim}$37,000  training examples on average, and the smallest language only has 1,794 examples. RCV1 and 2 are not easily accessible; a researcher who wishes to acquire the data would need to work with an organization that has obtained legal approval from Reuters Ltd. 

The XNLI dataset \cite{xnli} was designed for evaluating zero-shot cross-lingual transfer and does not contain training data for non-English languages.

The Yelp corpus \cite{yelp} contains reviews from international marketplaces, but the reviews from each marketplace can be written in multiple languages and the language identity is not provided. Furthermore, the Yelp corpus itself is refreshed from time to time, and previous versions are not made available for download, which affects the reproducibility of published results. 

Several versions of the Amazon reviews corpus exist today. Neither the version from \citet{ucsd_reviews} nor \citet{amazon} provide training, development, and test splits, and neither version focuses on the multilingual aspect of the reviews. \citet{prettenhofer-stein-2010-cross} provide Amazon reviews in 4 languages (i.e., 2,000 training and test reviews, along with a variable number of unlabeled reviews), but the dataset is small by modern standards.

We address many of the above-mentioned limitations by releasing a subset of Amazon reviews specifically tailored for the task of multilingual text classification:
\begin{itemize}
\itemsep0em 
    \item We provide 200,000 reviews in the training set for each of the languages in the corpus.
    \item We apply language detection algorithms to ensure reviews are associated with the correct language with high probability.
    \item We distribute the corpus on AWS Open Datasets for easy access by any research group for non-commercial purposes.
    \item Unlike previous Amazon reviews datasets, we split the data into clearly defined training, development, and test sets.
\end{itemize}
The Multilingual Amazon Reviews Corpus (MARC) can be found at \url{https://registry.opendata.aws/amazon-reviews-ml/}. The dataset description, code snippets, and license agreement can be retrieved at \url{https://docs.opendata.aws/amazon-reviews-ml/readme.html}.

\section{Data preparation}


\subsection{Inclusion Criteria}

We gathered the reviews from the marketplaces in the US, Japan, Germany, France, Spain, and China for the English, Japanese, German, French, Spanish, and Chinese languages, respectively. We considered reviews that were submitted between November 1, 2015 and November 1, 2019. Only reviews with verified purchases were included.

We take no more than 20 reviews from the same product, and no more than 20 reviews from the same reviewer. Only products with at least 2 reviews were included in the dataset. Reviews must be at least 20 characters long.

\subsection{Data Processing}

The language of a review does not necessarily match the language of its marketplace (e.g., reviews from Amazon.de are primarily written in German, but could also be written in English, etc.). For this reason, we applied a language detection algorithm \citep{fasttext} to determine the language of the review text. Only reviews written in the target language were retained. Based on a manual review of 200 randomly selected reviews per language, we observed 0, 0, 0, 0, 1, and 0 incorrectly classified reviews for English, Japanese, German, French, Spanish, and Chinese, respectively. At a score threshold of 0.8, the language filter removed 4.9\%, 0.2\%, 1.2\%, 2.4\%, 3.8\%, and 5.3\% of the English, Japanese, German, French, Spanish, and Chinese candidate reviews, respectively.

\begin{table*}[ht]
\begin{minipage}{1.0\linewidth}
	\centering
	\footnotesize
\begin{subtable}{1.0\linewidth}
\centering
\begin{tabu}{lccccccc}
\toprule
 & \textbf{En} & \textbf{De} & \textbf{Es} & \textbf{Fr} & \textbf{Ja} & \textbf{Zh} & \textbf{Average}\\
\midrule 
\textit{Fine-grained Classification} \\
\midrule 
Body only & 53.3 & 50.1 & 51.9 & 52.6 & 56.8 & 64.8 & 54.9 \\
Body, title \& category & 43.0 & 42.5 & 47.1 & 47.1 & 51.7 & 57.7 & 48.2 \\
\midrule
\textit{Binarized Classification} \\
\midrule 
Body only & 8.8 & 7.2 & 7.4 & 7.3 & 11.1 & 12.5 & 9.1 \\
Body, title \& category & 6.3 & 5.5 & 5.5 & 5.3 & 8.0 & 10.8 & 6.9 \\
\bottomrule
\end{tabu}
\caption{Fully supervised task (MAE$\times100$). The language of the training and test sets are the same.}
\label{table:mad_supervised}
\end{subtable}
\vspace{0.1cm}
\newline
\begin{subtable}{1.0\linewidth}
\centering
\begin{tabu}{cccccccc}
\toprule
\textbf{Source Lang.} & \textbf{En Test} & \textbf{De Test} & \textbf{Es Test} & \textbf{Fr Test} & \textbf{Ja Test} & \textbf{Zh Test} & \textbf{Average} \\
\midrule
En & - & 69.2 & 64.2 & 73.3 & 84.4 & 93.2 & 76.9 \\
De & 81.3 & - & 66.9 & 71.7 & 88.9 & 87.1 & 79.2 \\
Es & 73.6 & 68.4 & - & 65.7 & 92.5 & 85.2 & 77.1 \\
Fr & 77.5 & 68.4 & 61.7 & - & 88.6 & 86.4 & 76.5 \\
Ja & 78.5 & 77.6 & 71.5 & 82.4 & - & 83.8 & 78.8 \\
Zh & 78.8 & 77.9 & 79.1 & 84.1 & 84.3 & - & 80.8 \\
\bottomrule
\end{tabu}
\caption{Zero-shot cross-lingual transfer task (fine-grained classification MAE$\times100$). We train mBERT on source language data and test on non-source language data.}
\label{table:mad_zero-shot}
\end{subtable}
\vspace{0.3em}\newline
\begin{subtable}{1.0\linewidth}
\centering
\begin{tabu}{cccccccc}
\toprule
\textbf{Source Lang.} & \textbf{En Test} & \textbf{De Test} & \textbf{Es Test} & \textbf{Fr Test} & \textbf{Ja Test} & \textbf{Zh Test} & \textbf{Average} \\
\midrule
En & - & 15.5 & 10.7 & 14.0 & 19.4 & 27.6 & 17.4 \\
De & 15.3 & - & 11.4 & 14.6 & 23.6 & 21.6 & 17.3 \\
Es & 12.0 & 14.7 & - & 11.4 & 21.2 & 22.3 & 16.3 \\
Fr & 15.3 & 13.9 & 10.5 & - & 22.6 & 22.6 & 17.0 \\
Ja & 15.4 & 17.8 & 12.9 & 16.8 & - & 21.1 & 16.8 \\
Zh & 14.3 & 16.1 & 13.6 & 17.5 & 20.2 & - & 16.3 \\
\bottomrule
\end{tabu}
\caption{Zero-shot cross-lingual transfer task (binarized classification MAE$\times100$). We train mBERT on source language data and test on non-source language data.}
\label{table:mad_bin_zero-shot}
\end{subtable}
\vspace{0.3em}\newline
\begin{subtable}{1.0\linewidth}
\centering
\begin{tabu}{rccccccc}
\toprule
\textbf{En Train} & \textbf{En Test} & \textbf{De Test} & \textbf{Es Test} & \textbf{Fr Test} & \textbf{Ja Test} & \textbf{Zh Test} & \textbf{Non-En Average} \\
\midrule
6.25\% & 51.6 & 81.1 & 72.4 & 88.3 & 104.2 & 110.7 & 91.3 \\
12.5\% & 48.1 & 72.0 & 69.8 & 79.4 & 91.1 & 103.7 & 83.2 \\
25\% & 45.0 & 72.3 & 66.0 & 74.9 & 86.3 & 94.9 & 78.9 \\
50\% & 43.4 & 71.5 & 65.7 & 75.1 & 85.3 & 92.1 & 77.9 \\
100\% & 43.0 & 69.2 & 64.2 & 73.3 & 84.4 & 93.2 & 76.9 \\
\bottomrule
\end{tabu}
\caption{Amount of source language training data versus same-language and zero-shot transfer performance (fine-grained, MAE$\times100$). The training data comes from the English portion of the corpus only.}
\label{table:mad_fractions}
\end{subtable}
\end{minipage}
\caption{mBERT classification mean absolute error (MAE$\times100$). The `fine-grained' classification task predicts the 5-star rating, whereas the `binarized' task predicts whether the review is negative (i.e., 1-2 stars) or positive (i.e., 4-5 stars). Unless otherwise stated, we use the review body, review title, and product category as mBERT inputs.}
\label{table:mad}
\end{table*}
\begin{figure*}[h]
\centering
\includegraphics[width=16cm]{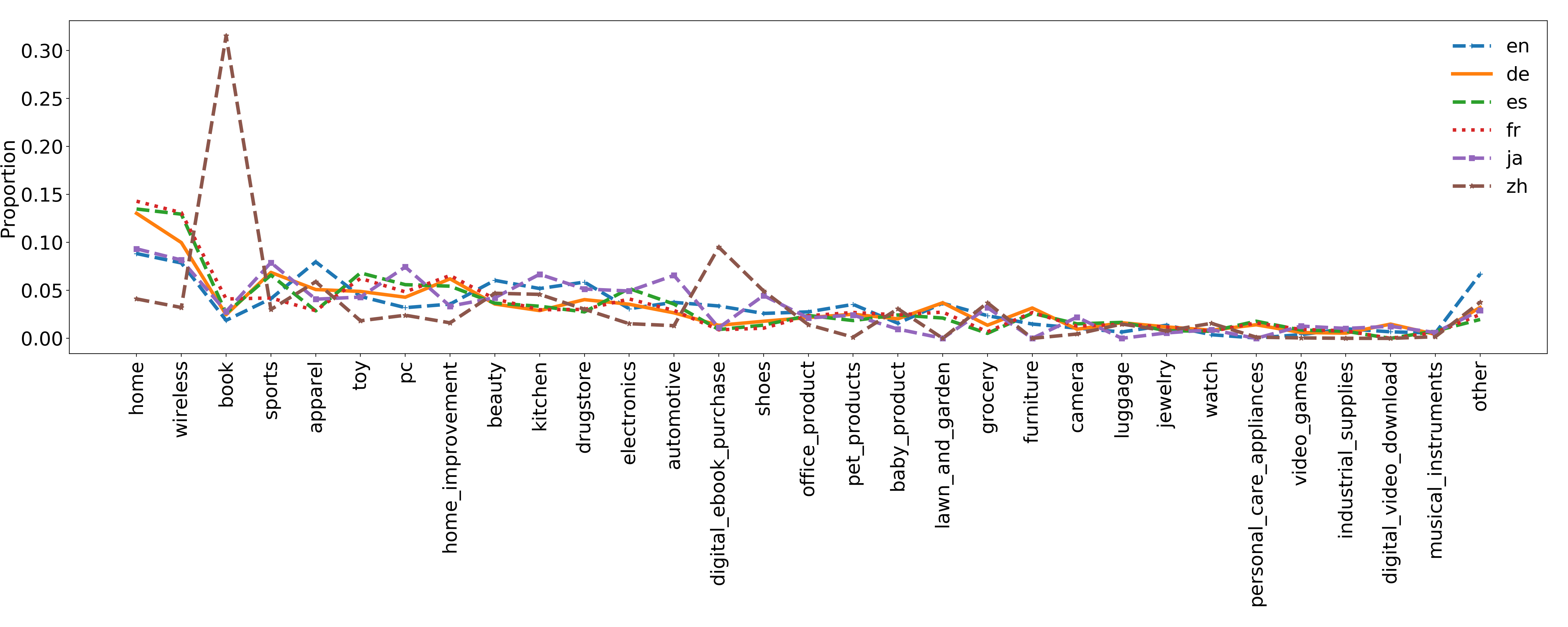}
\caption{Distribution of product categories across the training data in different languages.}
\label{fig:categories}
\end{figure*}

We also applied a vocabulary-based filter on the reviews. If a review contains a token that doesn't occur in at least 20 other reviews, then the review is excluded from the dataset. We used Jieba\footnote{\url{https://github.com/fxsjy/jieba}} for Chinese and KyTea\footnote{\url{http://www.phontron.com/kytea}} for Japanese word segmentation. The segmenters were only used during the filtering process, and the text provided in the dataset is not segmented or tokenized.

We truncate all reviews at 2,000 characters. Newlines and tabs in the body of the review were removed.

Some Amazon reviews contain HTML markup. We used Lynx\footnote{\url{https://lynx.invisible-island.net}} to render the reviews as UTF-8 plain-text.

Product and reviewer IDs were anonymized by mapping each ID to a unique randomly generated integer.

We provide the product category labels for 30 common product types, and all other product categories are mapped to `other'.

\section{Corpus Characteristics}

Amazon product ratings are given on a 5-star scale. To avoid any class imbalance issues in the dataset, we downsampled the reviews to ensure that each star rating constituted exactly 20\% of the corpus. We provide 200,000, 5,000, and 5,000 reviews for the training, development, and test sets, respectively.

In Table \ref{table:statistics}, we compile some of the important statistics for the corpus. The number of unique products and reviewers is broadly similar across different languages.

In Figure \ref{fig:categories}, we show the distribution of product categories for each language. There is substantial variation in the distribution of product categories by language. Chinese reviews, most notably, are heavily skewed towards books.

\section{Baseline Results}

In Table \ref{table:mad}, we provide baseline mean absolute error (MAE) results for supervised and zero-shot multilingual text classification with our corpus, where
\[ \mathit{MAE}(y,\hat{y}) = \frac{\sum_{i=1}^n |y_i - \hat{y}_i|}{n} \]
and $y_i, \hat{y}_i\in \{1,2,3,4,5\}$ are the true star rating and the predicted rating for the $i$-th review respectively. All of our baseline models are initialized with the cased multilingual BERT (mBERT) base model \cite{devlin2019bert}, which has 110M parameters.

Note that the star ratings for each review are \emph{ordinal}, and a 2-star prediction for a 5-star review should be penalized more heavily than a 4-star prediction for a 5-star review. However, previous work on Amazon reviews classification (e.g., \citealp{yang2016hierarchical}) used the classification accuracy as the primary metric, which ignores the ordinal nature of the labels. We use MAE in our baselines as the primary metric instead. We also report the classification accuracy for completeness (Table \ref{table:acc}), but we encourage the use of MAE in future work.

\subsection{Experimental Setup}

We predict the reviewer's rating using the text of the review (and possibly the product category) as the input. Following the procedure described in \citet{devlin2019bert}, we used the embedding of the CLS token for prediction. We fine-tuned the model for 15 epochs with the Adam optimizer using a constant learning rate of $8\times10^{-7}$. We used minibatches of 32 reviews. Each experiment required ${\sim}10$ hours to complete with a single GPU on an AWS p3.8xlarge instance with the MXNet GluonNLP framework. We truncated the review body at 180 wordpieces if it exceeded 180 wordpieces.

\subsection{Supervised Text Classification}

In Table \ref{table:mad_supervised}, we report our MAE on the fully supervised classification task, where the languages of the training and evaluation data are the same (i.e., train on French reviews and test on French reviews, etc.). We distinguish between the `fine-grained' classification task, where we predict on the 5-star scale, and the `binarized' classification task, where we predict whether the reviewer gave 1 to 2 stars or 4 to 5 stars. For the binarized task, we drop the 3-star reviews in the training and evaluation data.

We also distinguish between the case where the input is the body of the review alone and where the input is the review body combined with the review title and product category. In the latter case, we use mBERT for sentence pair classification, where the first `sentence' is the review body and the second `sentence' is the review title concatenated with the product category. The details for sentence pair classification can be found in \citet{devlin2019bert}.

\begin{table*}[ht]
\begin{minipage}{1.0\linewidth}
	\centering
	\footnotesize
\begin{subtable}{1.0\linewidth}
\centering
\begin{tabu}{lccccccc}
\toprule
 & \textbf{En} & \textbf{De} & \textbf{Es} & \textbf{Fr} & \textbf{Ja} & \textbf{Zh} & \textbf{Average}\\
\midrule 
\textit{Fine-grained Classification} \\
\midrule 
Body only & 56.5 & 58.3 & 56.9 & 55.5 & 53.9 & 51.4 & 54.4 \\
Body, title \& category & 63.3 & 62.0 & 58.9 & 58.8 & 57.2 & 55.1 & 59.2 \\
\bottomrule
\end{tabu}
\caption{Fully supervised task (classification accuracy). The language of the training and test sets are the same.}
\label{table:acc_supervised}
\end{subtable}
\vspace{0.1cm}
\newline
\begin{subtable}{1.0\linewidth}
\centering
\begin{tabu}{cccccccc}
\toprule
\textbf{Source Lang.} & \textbf{En Test} & \textbf{De Test} & \textbf{Es Test} & \textbf{Fr Test} & \textbf{Ja Test} & \textbf{Zh Test} & \textbf{Average} \\
\midrule
En  &   - &48.1&48.0&45.4&39.0&39.7&44.0 \\
De & 46.6&  - &47.9&46.9&38.9&40.0&44.1 \\
Es  & 48.8&47.5&  - &48.1&36.4&41.6&44.5 \\
Fr  & 48.1&47.1&49.5&  - &36.4&40.0&44.2 \\
Ja  & 45.2&41.5&45.2&39.6&  - &41.3&42.6 \\
Zh  & 44.6&43.2&43.5&41.8&40.0&  - &42.7 \\
\bottomrule
\end{tabu}
\caption{Zero-shot cross-lingual transfer task (fine-grained classification accuracy). We train mBERT on source language data and test on non-source language data.}
\label{table:acc_fine_zero-shot}
\end{subtable}

\vspace{0.3em}
\begin{subtable}{1.0\linewidth}
\centering
\begin{tabu}{rccccccc}
\toprule
\textbf{En Train} & \textbf{En Test} & \textbf{De Test} & \textbf{Es Test} & \textbf{Fr Test} & \textbf{Ja Test} & \textbf{Zh Test} & \textbf{Non-En Average} \\
\midrule
6.25\% & 59.1&44.2&45.5&40.7&35.8&36.6&40.5 \\
12.5\% & 61.7&46.5&45.6&45.0&37.8&38.3&42.6 \\
25\% & 62.9&46.6&47.6&45.5&38.1&38.5&43.2 \\
50\% & 63.2&47.4&47.7&45.6&38.6&39.9&43.9 \\
100\% & 63.3&48.1&48.0&45.4&39.0&39.7&44.0 \\
\bottomrule
\end{tabu}
\caption{Amount of source language training data versus same-language and zero-shot transfer performance (fine-grained, accuracy). The training data comes from the English portion of the corpus only.}
\label{table:acc_fractions}
\end{subtable}
\end{minipage}
\caption{mBERT classification accuracy. The `fine-grained' classification task predicts the 5-star rating. Unless otherwise stated, we use the review body, review title, and product category as mBERT inputs.}
\label{table:acc}
\end{table*}

\subsection{Zero-shot Text Classification}

In Tables \ref{table:mad_zero-shot} and \ref{table:mad_bin_zero-shot}, we report zero-shot cross-lingual transfer MAE for fine-grained and binarized classification respectively, where we only fine-tune mBERT on data from one source language and test the model on non-source languages. In our cross-lingual experiments, we used the review body, title, and product category as inputs.

Recent work by \citet{conneau2019unsupervised} recommended reporting zero-shot transfer results by using the target development sets for model checkpoint selection. In addition, \citet{keung_eval} showed that using the source language development set to select the checkpoint can lead to significant variation in zero-shot transfer performance and also recommended using the target development sets for checkpoint selection. Our results in Tables \ref{table:mad} and \ref{table:acc} follow their guidance, and we use the target development set to select the model checkpoint for each language.

In Table \ref{table:mad_fractions}, we vary the amount of English training data used in mBERT fine-tuning and examine the change in English test and non-English zero-shot MAE. Increasing the amount of English training data is generally helpful, although there are clearly diminishing returns.

\section{Conclusion}

We present a curated subset of Amazon reviews specifically designed to aid research in multilingual text classification. To the best of our knowledge, this is the largest public benchmark dataset for the training and evaluation of multilingual text classification models. With this work, we systematically address various gaps that we identified in existing multilingual corpora: we apply careful sampling, filtering, and text processing to the documents to minimize noise in the dataset, and we provide a large number of samples for training models in six languages with well-defined training, development, and test splits. We discuss the data preparation steps, analyze the distribution of the important characteristics of the corpus, and present baseline results for supervised and zero-shot cross-lingual text classification. With these contributions, we hope that this corpus will be an important resource to the research community.

\section{Acknowledgments}

We thank the anonymous reviewers for their insightful feedback. We also thank Peter Schmiedeskamp and Sharon Chiarella for their support in the preparation and release of this corpus. NAS acknowledges NSF support under grant 1813153.

\bibliography{anthology,emnlp2020}
\bibliographystyle{acl_natbib}

\end{document}